\documentclass[10pt,twocolumn,letterpaper]{article}

\usepackage{cvpr}              % To produce the CAMERA-READY version
\usepackage{multirow}
\usepackage{subcaption}
\usepackage{graphicx}
\usepackage{booktabs} 
\usepackage{amssymb}
\usepackage{xcolor}
\usepackage{colortbl}
\usepackage[table,xcdraw]{xcolor}
\usepackage{tabularx}
\usepackage{amsmath}
\definecolor{cvprblue}{rgb}{0.21,0.49,0.74}
\usepackage[pagebackref,breaklinks,colorlinks,allcolors=cvprblue]{hyperref}

\def\paperID{17901} % *** Enter the Paper ID here
\def\confName{CVPR}
\def\confYear{2026}

\title{Unify the Views: View-Consistent Prototype Learning for Few-Shot Segmentation}

\author{
Hongli Liu$^{1}$, Yu Wang$^{1}$\thanks{Corresponding author.}, Shengjie Zhao$^{1*}$\\[0.6em]
$^{1}$School of Computer Science and Technology, Tongji University\\[0.4em]
\texttt{hongli01@tongji.edu.cn, yuwangtj@yeah.net, shengjiezhao@tongji.edu.cn}\\[0.2em]
}

\begin{document}
\maketitle
\begin{abstract}
Few-shot segmentation (FSS) has gained significant attention for its ability to generalize to novel classes with limited supervision, yet remains challenged by structural misalignment and cross-view inconsistency under large appearance or viewpoint variations. This paper tackles these challenges by introducing VINE (View-Informed NEtwork), a unified framework that jointly models structural consistency and foreground discrimination to refine class-specific prototypes. Specifically, VINE introduces a spatial–view graph on backbone features, where the spatial graph captures local geometric topology and the view graph connects features from different perspectives to propagate view-invariant structural semantics. To further alleviate foreground ambiguity, we derive a discriminative prior from the support-query feature discrepancy to capture category-specific contrast, which reweights SAM features by emphasizing salient regions and recalibrates backbone activations for improved structural focus. The foreground-enhanced SAM features and structurally enriched ResNet features are progressively integrated through masked cross-attention, yielding class-consistent prototypes used as adaptive prompts for the SAM decoder to generate accurate masks. Extensive experiments on multiple FSS benchmarks validate the effectiveness and robustness of VINE, particularly under challenging scenarios with viewpoint shifts and complex structures.  
The code is available at \url{https://github.com/HongliLiu1/VINE-main}.
\end{abstract}    
\section{Introduction}
\label{sec:intro}

Semantic segmentation assigns category-level labels to every pixel and plays a foundational role in visual scene understanding~\cite{long2015fully,strudel2021segmenter,cheng2022masked}.
Nevertheless, fully-supervised segmentation methods rely on large-scale pixel-wise annotations, which are costly to obtain and difficult to scale, particularly for long-tailed or open-set scenarios. 
Few-Shot Segmentation (FSS)~\cite{shaban2017one,liu2025devil,peng2023hierarchical,wang2023rethinking,tong2025self} addresses this limitation by enabling segmentation of novel categories with only a few labeled examples. 
By leveraging support-query interactions to transfer knowledge from base to novel classes, FSS establishes a label-efficient paradigm that generalizes beyond annotated categories.

\begin{figure}[t]
    \centering
    \includegraphics[width=\linewidth]{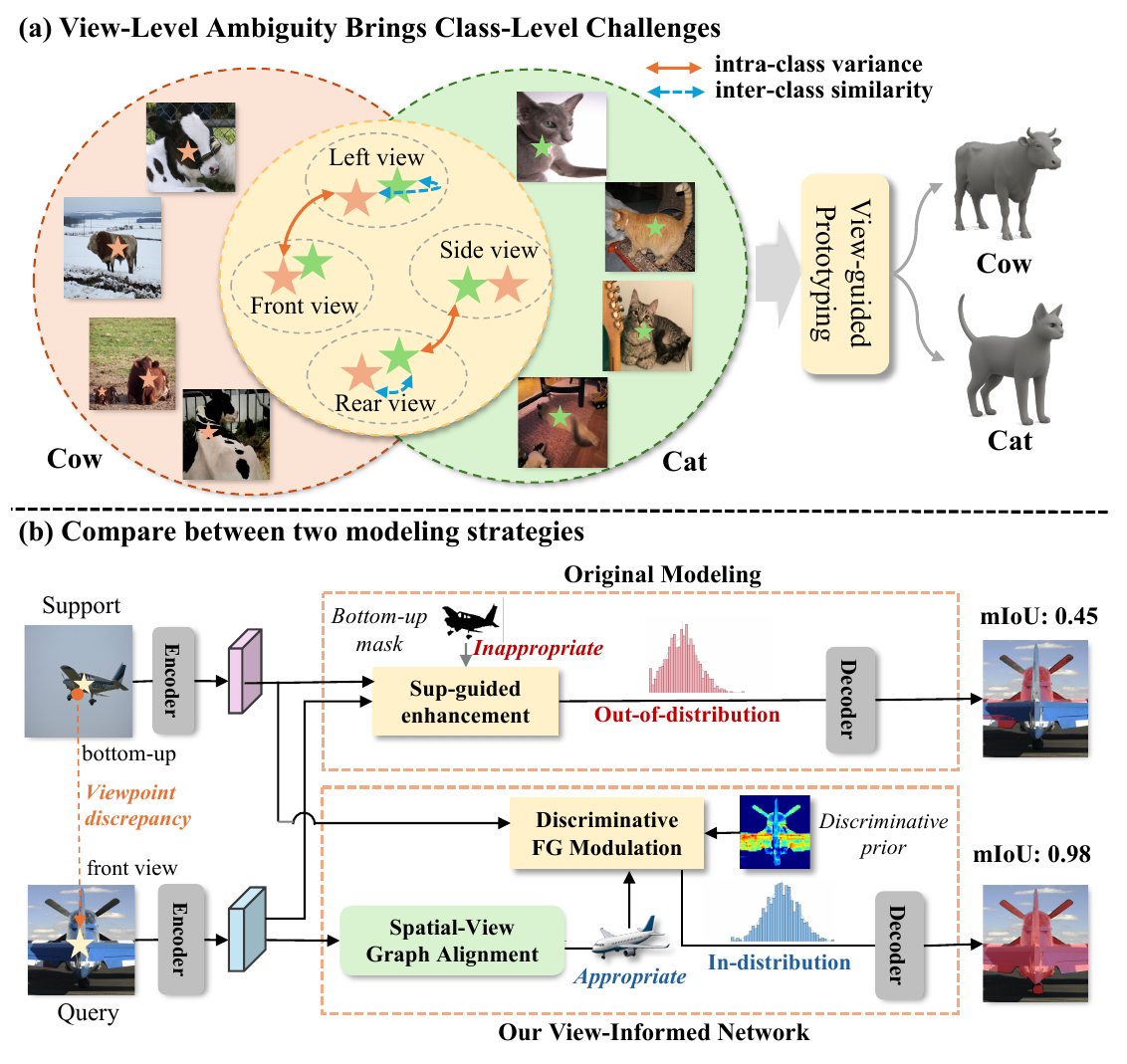}
    \caption{Comparison between conventional prototype modeling and VINE. 
\textbf{(a)} View-induced challenges: large intra-class variation and inter-class similarity lead to prototype confusion. 
\textbf{(b)} Top: conventional mask-guided enhancement fails under viewpoint shifts, generating unstable prototypes. Bottom: VINE unifies spatial–view graph alignment and discriminative modulation to produce view-consistent, structurally reliable prototypes.}
    \label{fig:motivation}
    \vspace{-1.0em}
\end{figure}
Most few-shot segmentation (FSS) methods follow a prototype paradigm: class representations are distilled from support features and matched against the query to drive dense prediction. Early systems~\cite{wang2019panet,tian2020prior} form prototypes via global average pooling and segment by feature similarity. Subsequent work replaces global pooling with cross-attention~\cite{boudiaf2021few,shi2022dense,min2021hypercorrelation,wang2023rethinking} to model pixel-level support–query correspondences and inject support cues into the query stream. 
Despite these advances, Fig.~\ref{fig:motivation}(a) reveals a persistent failure mode: cross-view geometric variation and fine-grained inter-class resemblance (e.g., ``Cat'' vs.\ ``Cow'') often destabilize correspondence, leading to prototype drift and boundary ambiguity. 
In essence, structural consistency, which ensures part–whole geometric preservation across viewpoints, remains insufficiently constrained.
Recent pipelines have turned to foundation models, notably SAM~\cite{kirillov2023segment,zhang2024bridge}, which offer prompt-driven mask generation as a source of generalizable cues. Despite these advances, as illustrated in Fig.~\ref{fig:motivation}(b), SAM’s instance-centric saliency prior and sensitivity to prompt placement do not transfer reliably in FSS, where support and query depict different instances with substantial pose and shape variation. As a result, pseudo-mask guidance often propagates inconsistently across mismatched poses (e.g., lateral to frontal views), imposing incorrect structural cues, while the lack of explicit foreground–background modeling makes the system vulnerable to noisy masks~\cite{xu2024eliminating} and occlusion. These observations naturally raise a key question: how can we explicitly couple cross-view structural alignment with robust, query-adaptive foreground discrimination?

To address these challenges, we propose VINE (View-Informed Network), a unified framework that jointly enforces structural consistency and semantic discrimination for robust prototype learning under cross-view variations. We posit that high-quality visual prompts should maintain spatial alignment and class consistency while suppressing background noise.
VINE adopts a dual-encoder architecture that combines semantic-rich representations from the frozen SAM encoder and structure-sensitive features from ResNet. It integrates two complementary modules: the Spatial-View Graph Alignment (SVGA) module constructs spatial–view graphs over ResNet features to model intra-class geometry across viewpoints, where a prototype loss further enforces alignment between query features and view-modeled prototypes; the Discriminative Foreground Modulation (DFM) mechanism generates class-consistent priors through prototype–query interaction, enhancing discriminative focus and suppressing distractors.
As illustrated in Fig.~\ref{fig:motivation}(b), by decoupling geometry and semantics and guiding query features with appropriate priors, VINE produces in-distribution prototypes with improved cross-view generalization. Finally, learnable tokens encode aggregated foreground and class-consistent cues via cross-attention with SAM and ResNet features, and the fused prototypes serve as prompts for SAM’s decoder to generate final masks.

\vspace{0.5em}
\noindent{In summary, our contributions are as follows:}
\begin{itemize}
  \item We propose VINE, a unified framework that enhances robust few-shot segmentation by jointly modeling structural consistency and foreground discrimination.
  \item We design Spatial-View Graph Alignment to build spatial–view graphs that capture intra-class structure and cross-view consistency, with a prototype loss enforcing prototype-level correspondence across views.
  \item We introduce Discriminative Foreground Modulation, leveraging prototype-query interaction to generate foreground-aware priors for improved discriminability.
  \item We demonstrate the effectiveness of the proposed view-informed framework through extensive experiments on standard few-shot segmentation benchmarks with diverse viewpoints and complex scenes.
\end{itemize}
\section{Related Work}
\label{sec:relatedwork}

\subsection{Metric-Based FSS}
Few-Shot Segmentation (FSS) aims to segment novel-class objects in query images using only a few labeled support examples. Metric-based approaches address this by extracting class prototypes from support features and matching them with query features in a shared embedding space~\cite{shaban2017one,tian2020prior,wang2023rethinking,wang2024adaptive}. Early methods typically compute a single or few global prototypes and perform pixel-wise similarity matching, which often overlooks spatial structure and fails under significant appearance variation.
To enhance support–query alignment, attention-based methods~\cite{zhang2019pyramid,xu2023self,liu2023fecanet,park2024task,boudiaf2021few,min2021hypercorrelation,zhang2021few} incorporate fine-grained interactions by dynamically fusing each query pixel with dense foreground features from the support set to emphasize semantically relevant regions.
Recent efforts incorporate structure-aware representations~\cite{liu2023few,schwingshackl2025few} to model contextual dependencies via part-level reasoning, instance associations. 
These methods often overlook intra-class variation and lack mechanisms to enforce cross-view consistency, leading to suboptimal alignment under large appearance and geometric changes.
% 跨栏顶部插图
\begin{figure*}[t]
  \centering
  \includegraphics[width=0.89\textwidth]{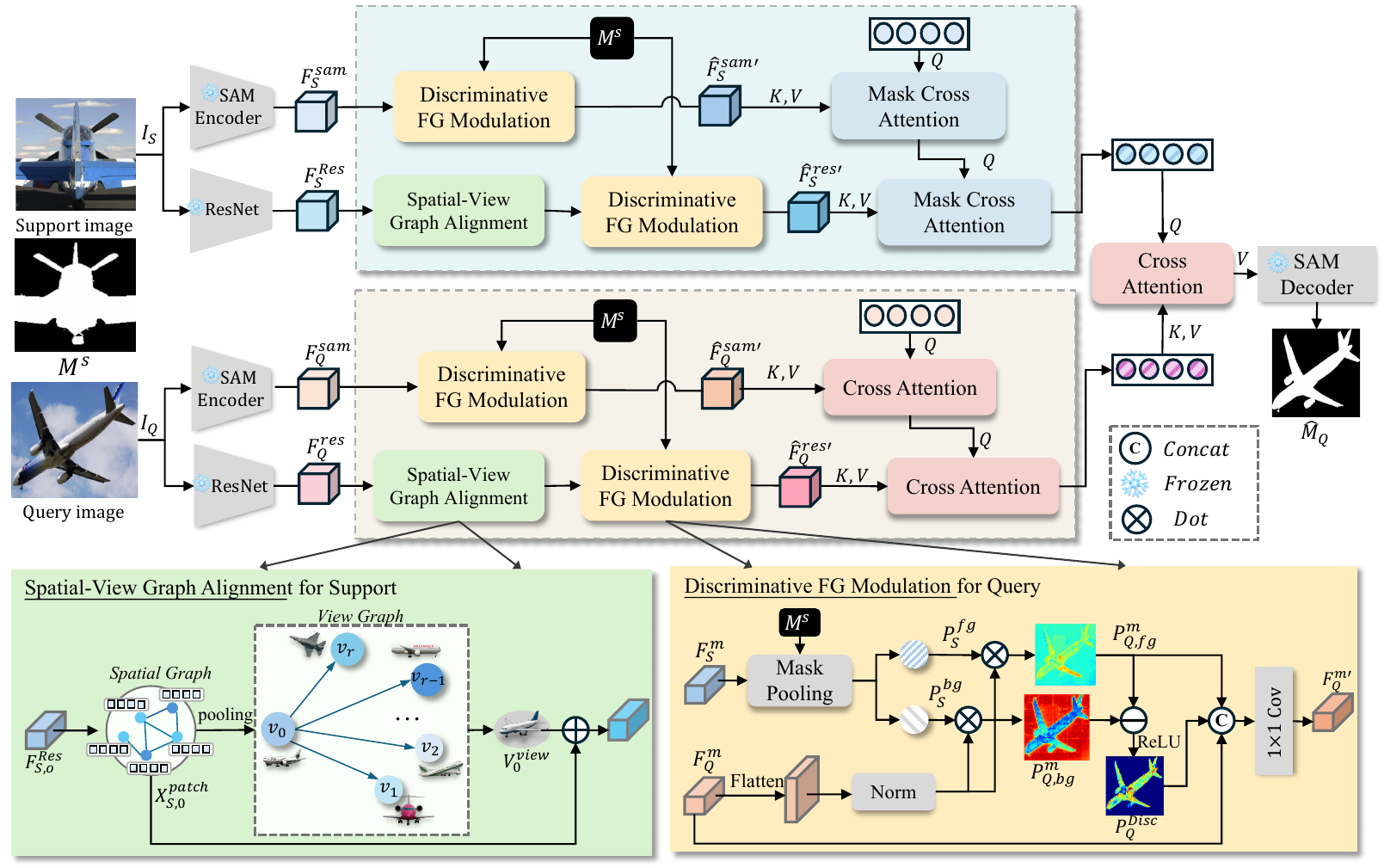}
  \caption{An overview of our proposed framework. We incorporate Spatial-View Graph Alignment (SVGA) to capture structural consistency across views, and Discriminative Foreground Modulation (DFM) to highlight foreground-relevant features, collaboratively improving the quality of support-query prototypes and guiding accurate mask prediction.}
  \label{fig:overview}
  \vspace{-1em} % 调整图像和正文之间的垂直间距
\end{figure*}
\subsection{Foundation Model-Based FSS}
Recent advances have integrated foundation models, particularly the Segment Anything Model (SAM)\cite{kirillov2023segment}, into Few-Shot Segmentation (FSS) to enhance generalization and reduce annotation costs\cite{liu2023matcher,wang2023samrs,he2024apseg,meng2024segic,liu2024simple}. SAM provides strong class-agnostic segmentation via prompt-based guidance, making it suitable for low-supervision settings. Beyond SAM-based pipelines, HMNet~\cite{xu2024hybrid} employs a hybrid Mamba architecture to strengthen long-range dependency modeling for prototype construction, while UniFSS~\cite{chang2025beyond} redefines guidance by unifying cues beyond masks to improve supervision. Existing SAM-enhanced approaches such as VRP-SAM~\cite{sun2024vrp} and SAM-RS~\cite{xu2024sam} improve prompt quality through visual-reference encoding and multimodal refinement, and FCP~\cite{park2025foreground} adopts prototype-guided prompting to align support–query semantics. However, these methods largely rely on appearance similarity or prompt refinement, lacking explicit mechanisms to preserve structural coherence and foreground–background separation under geometric variation. Therefore, coupling structural alignment with discriminative foreground modeling is crucial for reliable and generalizable visual prompts in FSS.
\section{Method}
\label{sec:method}
\subsection{Preliminaries \& Overview}
\paragraph{Preliminaries.}
Few-Shot Segmentation (FSS) targets pixel-level segmentation of novel classes using only a few annotated examples. It is typically formulated within a meta-learning framework, with training and testing performed on disjoint datasets $\mathcal{D}_{\text{train}}$ and $\mathcal{D}_{\text{test}}$, whose label spaces $\mathcal{C}_{\text{base}}$ and $\mathcal{C}_{\text{novel}}$ satisfy $\mathcal{C}_{\text{base}} \cap \mathcal{C}_{\text{novel}} = \emptyset$.
Following the episodic training paradigm, each episode consists of a support set $\mathcal{S} = \{(I_i^S, M_i^S)\}_{i=1}^K$ and a query pair $(I^Q, M^Q)$, where $I$ denotes the image and $M$ is its binary mask 
% (1 for foreground, 0 for background). 
The support and query images belong to the same class, but $M^Q$ is not revealed during inference.
% During training, the model learns to predict $\hat{M}^Q$ for $I^Q$ using $\mathcal{S}$, supervised by classes in $\mathcal{C}_{\text{base}}$. At test time, the learned segmentation is transferred to $\mathcal{C}_{\text{novel}}$. 
We adopt the standard $k$-shot setting and default to the 1-shot case, which can be extended to $k$-shot via feature aggregation.
\paragraph{Overview.}
The overall pipeline of VINE is illustrated in Fig.~\ref{fig:overview}.
Given a support–query pair $(I_S, I_Q)$, we extract semantic and structural features ${F_S^m, F_Q^m}$ from the frozen SAM encoder and the ResNet backbone, where $m \in \{sam, res\}$. 
To enhance cross-view structural consistency, the Spatial-View Graph Alignment (SVGA) module refines the structural features via spatial–view graph propagation, yielding $(F_S^{res’}, F_Q^{res’})$.
The semantic features and refined structural features are then processed by the Discriminative Foreground Modulation (DFM) mechanism, which leverages support masks and support–query interactions to highlight foreground regions, producing modulated features $(F_S^{m^\prime}, F_Q^{m^\prime})$ for $m \in \{sam, res^\prime\}$.
A set of learnable prototype tokens first interact with modulated features via Masked Cross-Attention to obtain the support-aware prototype $P_S$, and subsequently through Cross-Attention to derive the query-adaptive prototype $P_Q$. A final Cross-Attention fuses $P_S$ and $P_Q$ into a view-informed Visual Reference Prompt (VRP), which is fed into the SAM decoder to produce the final mask $\hat{M}_Q$.
\subsection{Spatial-View Graph Alignment}
Motivated by the need to preserve geometric coherence across varying viewpoints, the Spatial–View Graph Alignment (SVGA) module models intra-class consistency by coupling spatial and view-level graphs. 
This dual-graph formulation captures geometric relations within instances while enforcing cross-view alignment between support and query representations, yielding structure-aware and view-consistent features.

Limited viewpoint diversity in the 1-shot setting motivates the generation of $R$ perturbed support views through homography-based transformations:
\begin{align}
\tilde{I}_S^{r} &= \text{Warp}(I_S, H^{r}), \
H^{r} &= \text{PerturbCorners}(\delta_{\max}),
\end{align}
where $H^{r} \in \mathbb{R}^{3 \times 3}$ denotes the homography matrix obtained by corner perturbation within $\delta_{\max}$, and $\text{Warp}(\cdot)$ applies bilinear warping under $H^r$. 

Each augmented support view $\tilde{I}_S^r$ is then encoded by a shared backbone to produce structural feature maps $\tilde{F}_S^{\text{res},r}$, while the query image $I_Q$ is processed in the same manner to yield $F_Q^{\text{res}} \in \mathbb{R}^{C \times H \times W}$. 
All features are reshaped into $N = H \times W$ patch embeddings $\mathbf{X}_{S,r},\ \mathbf{X}_Q \in \mathbb{R}^{N \times C}$, serving as nodes for subsequent spatial graph construction.

For spatial modeling, a graph $\mathcal{G}_{\text{space}} = (\mathbf{X}_*, \mathcal{E}_{\text{space}})$ is constructed over the patch embeddings $\mathbf{X}_*$, where $* \in \{S, Q\}$ and $r=0$ denotes either the original support view or the query image. Each node corresponds to a patch token located at coordinate $\mathbf{p}_j$ on the feature grid, and the edge set $\mathcal{E}_{\text{space}} = \text{KNN}(\{\mathbf{p}_j\}_{j=1}^{N}, k)$ is defined by connecting each node to its $k$ nearest neighbors in Euclidean space, thereby capturing spatial adjacency.

A Graph Attention Network (GAT)~\cite{velickovic2017graph} is then applied over $\mathcal{G}_{\text{space}}$ to aggregate contextual cues among spatially adjacent patches rather than individual pixels, effectively modeling local geometric relations while maintaining computational efficiency. This yields context-aware patch embeddings $\mathbf{X}_*^{\text{patch}}$ for subsequent view-level modeling:
\begin{equation}
\mathbf{X}_*^{\text{patch}} = \text{GAT}_{\text{space}}(\mathbf{X}_*, \mathcal{E}_{\text{space}}),
\end{equation}
producing context-aware patch embeddings $\mathbf{X}_*^{\text{patch}}$ for subsequent view-level modeling.

Beyond spatial modeling, a view-level graph $\mathcal{G}_{\text{view}}=(\{\mathbf{v}_r\},\mathcal{E}_{\text{view}})$ is constructed to capture viewpoint-induced variability and to enforce semantic coherence across support views. For each support view $r\!\in\!\{0,\dots,R\}$, a node embedding is obtained by average pooling over spatial patches, $\mathbf{v}_r=\frac{1}{N}\sum_{j=1}^{N}\mathbf{X}^{\text{patch}}_{S,r,j}\!\in\!\mathbb{R}^C$, which summarizes view-level semantics while being robust to local deformations. 
To anchor all auxiliary views to a canonical reference and avoid spurious pairwise cycles, edges adopt a star topology $\mathcal{E}_{\text{view}}=\{(0,r)\mid r=1,\dots,R\}$ with the original view ($r=0$) as hub; this design stabilizes message passing, reduces over-smoothing from dense connectivity, and preserves a consistent alignment space. 

A view-level GAT is applied over $\mathcal{G}_{\text{view}}$ to propagate semantic cues across views and aggregate complementary information from multiple perspectives, reinforcing structural consistency among them. Through attention-based message passing, each node embedding $\mathbf{v}_r^{\text{view}}$ integrates context from related views while preserving canonical semantics:
\begin{equation}
\mathbf{v}_r^{\text{view}} = \text{GAT}_{\text{view}}({\mathbf{v}_j}, \mathcal{E}_{\text{view}}).
\end{equation}

Original-support patch features are refined by linearly combining local spatial embeddings with the view-aggregated global representation:
\begin{equation}
\mathbf{X}_{S,0}^{\text{SVGA}} = \mathbf{X}_{S,0}^{\text{patch}} + \mathbf{v}_{0}^{\text{view}},
\end{equation}
where $\mathbf{v}_{0}^{\text{view}}$ is broadcast over $N$ positions to match $\mathbf{X}_{S,0}^{\text{patch}}$, enabling local–global integration.

In the 1-shot case, query augmentation is omitted to avoid semantic distortion under limited supervision, restricting view-level modeling to the support branch; 
For $K$-shot ($K>1$), both support and query branches contain multiple real samples across diverse viewpoints, allowing semantic information exchange and cross-view coherence through a fully connected view graph within each branch. 
Finally, the patch features $\mathbf{X}_{Q/S}^{\text{SVGA}}$ are reshaped into $F_{Q/S}^{\text{Res}’}$ for prototype matching.

To enforce structural alignment across views, we introduce a prototype consistency constraint over the SVGA-enhanced representations. 
Specifically, we perform global average pooling over the fused spatial features to derive class-level prototypes from the support and query branches:
\begin{equation}
\mathbf{p}_v^{S} = \text{GAP}(F_{S,0}^{\text{Res}'}), \quad \mathbf{p}_v^{Q} = \text{GAP}(F_Q^{\text{Res}'}),
\end{equation}
where $\mathbf{p}_v^{S}, \mathbf{p}_v^{Q} \in \mathbb{R}^C$ capture the global structure-aware semantics of each modality.

The prototype consistency loss $\mathcal{L}_{\text{proto}}$ is defined as the mean squared error (MSE) between the class-wise prototypes to ensure feature alignment:
\begin{equation}
\mathcal{L}_{\text{proto}} = \text{MSE}(\mathbf{p}_v^{Q}, \mathbf{p}_v^{S}),
\end{equation}
which minimizes the Euclidean distance between support and query prototypes to enforce cross-view structural coherence in the prototype space.

\subsection{Discriminative Foreground Modulation}

Relying solely on pixel-level similarity is insufficient to capture category-specific foreground cues in few-shot segmentation, especially under appearance shifts. Inspired by this limitation, the Discriminative Foreground Modulation (DFM) mechanism is introduced to incorporate semantic–structural priors and mask supervision, enhancing the model’s ability to emphasize task-relevant regions.
We begin by extracting class-specific foreground and background prototypes from both structural and semantic modalities:
\begin{equation}
\begin{aligned}
P_{Q,fg}^m &= \cos(F_Q^m, P_S^{\text{fg}}),\quad
P_{Q,bg}^m &= \cos(F_Q^m, P_S^{\text{bg}}),\\
\end{aligned}
\label{eq:disc_prior}
\end{equation}
where $m \in \{\text{res}', \text{sam}\}$ denotes the modality, $\odot$ is element-wise multiplication, and $\text{GAP}(\cdot)$ performs global average pooling over spatial dimensions.

Foreground–background distinction in the query image is guided by semantic relevance computed with cosine similarity between query and support-derived prototypes:
\begin{equation}
P_{Q,fg}^m = \cos(F_Q^m, P_S^{\text{fg}}), \quad
P_{Q,bg}^m = \cos(F_Q^m, P_S^{\text{bg}}), 
\end{equation}
\begin{equation}
  P_Q^{\text{Disc}} = \text{ReLU}(P_Q^{\text{fg}} - P_Q^{\text{bg}}),
\end{equation}
where $P_{Q,fg}^m$ and $P_{Q,bg}^m$ measure the semantic correspondence of query features to the support-derived foreground and background prototypes, respectively. Their contrast yields the discriminative prior $P_Q^{\text{Disc}}$, which highlights foreground-dominant regions and suppresses background responses, providing fine-grained cues for subsequent modulation and prototype refinement.

Semantic cues are then incorporated into both branches for feature refinement.
For the support branch, we concatenate the SVGA-enhanced feature $F_S^m$, the semantic prototype $P_S^{\text{fg}}$, and the binary mask $M_S$ along the channel dimension, followed by a $1 \times 1$ convolution:
\begin{equation}
F_S^{m’} = \text{Conv}_{1 \times 1}(\text{Concat}(F_S^m,\ P_S^{\text{fg}},\ M_S)).
\end{equation}
Where $\text{Concat}(\cdot)$ denotes channel-wise concatenation.

For the query branch, we analogously integrate $F_Q^m$, $P_S^{\text{fg}}$, and the discriminative $P_Q^{\text{Disc}}$ to enforce semantic refinement:
\begin{equation}
F_Q^{m'} = \text{Conv}_{1 \times 1}(\text{Concat}(F_Q^m,\ P_S^{\text{fg}},\ P_Q^{\text{Disc}})).
\end{equation}
The modulated features $F_S^{m'}$ and $F_Q^{m'}$ are used for prototype refinement and final mask prediction.

\subsection{Visual Reference Prompt Generation}
Reliable cross-view guidance is realized through a visual reference prompt generation module that constructs structure-consistent and class-discriminative prototypes for precise mask prediction. The module integrates semantic and geometric cues from both support and query via masked and standard cross-attention, yielding refined prompts that guide the SAM decoder toward accurate segmentation.

Two learnable prompt tokens, $\mathbf{P}^S_0$ and $\mathbf{P}^Q_0 \in \mathbb{R}^{1 \times C}$, are initialized for the support and query inputs. 
The support prompt $\mathbf{P}^S_0$ first attends to the semantic feature $F_S^{\text{sam}’}$ under mask guidance to capture task-relevant cues, and then refines its representation by attending to the structural feature $F_S^{\text{res}’}$ while keeping $F_S^{\text{sam}’}$ as the key:
\vspace{-0.1em}
\begin{equation}
\begin{aligned}
\mathbf{P}^S_1 &= \text{MaskedCrossAttn}(\mathbf{P}^S_0,\ F_S^{\text{sam}'},\ F_S^{\text{sam}'},\ M_S),\\
\mathbf{P}^S &= \text{MaskedCrossAttn}(\mathbf{P}^S_1,\ F_S^{\text{sam}'},\ F_S^{\text{res}'},\ M_S).
\end{aligned}
\end{equation}
the two-stage attention allows $\mathbf{P}^S$ to encode semantic relevance and geometric structure for cross-branch prompting.

Similarly, the query prompt $\mathbf{P}^Q_0$ undergoes progressive refinement through dual cross-attention over its SAM and ResNet features, yielding $\mathbf{P}^Q_1$ and $\mathbf{P}^Q$ that jointly encode semantic relevance and spatial structure:
\begin{equation}
\begin{aligned}
\mathbf{P}^Q_1 &= \text{CrossAttn}(\mathbf{P}^Q_0,\ F_Q^{\text{sam}'},\ F_Q^{\text{sam}'}), \\
\mathbf{P}^Q &= \text{CrossAttn}(\mathbf{P}^Q_1,\ F_Q^{\text{sam}'},\ F_Q^{\text{res}'}).
\end{aligned}
\end{equation}

Cross-attention bridges $\mathbf{P}^S$ and $\mathbf{P}^Q$ to achieve semantic–structural correspondence, aligning class-level context from the support with instance-specific cues from the query, serving as the unified visual reference prompt:
\begin{equation}
\mathbf{P}^{\text{VRP}} = \text{CrossAttn}(\mathbf{P}^S,\ \mathbf{P}^Q,\ \mathbf{P}^Q).
\end{equation}

The resulting visual reference prompt $\mathbf{P}^{\text{VRP}}$ serves as a mask prompt, injecting cross-view semantic and structural priors into the SAM decoder for precise mask generation:
\begin{equation}
\hat{M}_Q = \text{SAMDecoder}(\mathbf{P}^{\text{VRP}}),
\end{equation}
where $\hat{M}_Q \in [0,1]^{H \times W}$ is supervised by a hybrid loss that balances pixel-level precision and region-level overlap:
\begin{equation}
\mathcal{L}_{\text{Pred}} = \mathcal{L}_{\text{BCE}}(\hat{M}_Q, M^Q) + \mathcal{L}_{\text{DL}}(\hat{M}_Q, M^Q),
\end{equation}
where $\mathcal{L}_{\text{BCE}}$ ensures pixel-wise precision by penalizing misclassified foreground/background pixels, while $\mathcal{L}_{\text{DL}}$ (Dice Loss) emphasizes region-level overlap to mitigate class imbalance. $M^Q$ denotes the ground-truth mask.

The overall objective unifies structural alignment and mask prediction into a single optimization framework, combining the prototype consistency loss for cross-view coherence with the mask prediction loss for spatial accuracy:
\begin{equation}
\mathcal{L}_{\text{total}} = \lambda_{\text{proto}} \mathcal{L}_{\text{proto}} + \lambda_{\text{Pred}} \mathcal{L}_{\text{Pred}}.
\end{equation}

\begin{table*}[t]
\centering
\footnotesize   % 或改为 \footnotesize, \scriptsize
\normalsize
\setlength{\tabcolsep}{2pt} 
\renewcommand{\arraystretch}{1.2}  % 增加行高
\resizebox{0.9\textwidth}{!}{
\begin{tabular}{c|c|ccccc|ccccc|ccccc|ccccc}
\hline
\multirow{3}{*}{\centering\textbf{Method}} & \multirow{3}{*}{\centering\textbf{IE}} 
& \multicolumn{10}{c|}{\textbf{PASCAL-5$^i$}} 
& \multicolumn{10}{c}{\textbf{COCO-20$^i$}} \\
\cline{3-12} \cline{13-22}
& &
\multicolumn{5}{c|}{\textbf{1-shot}} & \multicolumn{5}{c|}{\textbf{5-shot}} 
& \multicolumn{5}{c|}{\textbf{1-shot}} & \multicolumn{5}{c}{\textbf{5-shot}} \\
\cline{3-12} \cline{13-22}
& & F-0 & F-1 &F-2& F-3 & Mean 
  & F-0 & F-1 & F-2 & F-3 & Mean 
  & F-0 & F-1 & F-2 & F-3 & Mean 
  & F-0 & F-1 & F-2 & F-3 & Mean \\
\hline
HDMNet~\cite{peng2023hierarchical}         & \multirow{5}{*}{\rotatebox{90}{VGG}}
              & 64.8 & 71.4 & 67.7 & 56.4 & 65.1
              & 68.1 & 73.1 & 71.8 & 64.0 & 69.3
              & 40.7 & 50.6 & 48.2 & 44.0 & 45.9
              & 47.0 & 56.5 & 54.1 & 51.9 & 52.4 \\
VRP-SAM$^\dag$~\cite{sun2024vrp} & 
              & 69.9 & 73.9 & 67.6 & 62.3 & 68.4
              & 72.8 & 75.0 & 67.5 & 63.2 & 69.6
              & 39.4 & 52.0 & 50.6 & 47.5 & 47.4
              & 43.6 & 57.7 & 54.7 & 51.8 & 51.9 \\
HMNet~\cite{xu2024hybrid} &
              &66.7 & 74.5 & 68.9 & 59.0 & 67.3 
              &70.5 & 76.0 & 72.2 & 65.7 & 71.1 
              &44.2 & 51.8 & 51.9 & 48.4 & 49.1 
              &48.8 & 58.0 & 57.9 & 53.2 & 54.5 \\
FCP$^\dag$~\cite{park2025foreground}      & 
              & 71.9 & 75.0 & 69.8 & 65.5 & 70.5
              & 73.5 & 76.4 & 71.6 & 65.8 & 71.8
              & 44.5 & 54.3 & 53.5 & 47.7 & 50.0
              & 49.0 & 58.3 & 55.2 & 52.3 & 53.7 \\
\rowcolor{gray!15}
VINE (Ours)     & 
              & \textbf{72.6} & \textbf{75.9} & \textbf{71.5} & \textbf{65.8} & \textbf{71.4}
              & \textbf{74.0} & \textbf{77.2} & \textbf{72.8} & \textbf{66.3} & \textbf{72.6}
              & \textbf{45.2} & \textbf{55.4} & \textbf{54.9} & \textbf{48.6} & \textbf{51.0}
              & \textbf{49.8} & \textbf{59.4} & \textbf{56.1} & \textbf{52.7} & \textbf{54.5} \\
\hline
PFENet~\cite{tian2020prior}         & \multirow{10}{*}{\rotatebox{90}{ResNet}}
              &61.7&69.5& 55.4&56.3&60.8              & 63.1 & 70.7 & 55.8 &57.9& 61.9
              & 36.5 & 38.6 & 34.5 & 33.8 & 35.8
              & 36.5 & 43.3 & 37.8 & 38.4 & 39.0 \\
CyCTR~\cite{zhang2021few}          & 
              & 65.7 & 71.0 & 59.5 & 59.7 & 64.0
              & 69.3 & 73.5 & 63.8 & 63.5 & 67.5
              & 38.9 & 43.0 & 39.6 & 39.8 & 40.3
              & 41.1 & 48.9 & 45.2 & 47.0 & 45.6 \\
SSP~\cite{fan2022self}            & 
              & 60.5 & 67.8 & 66.4 & 51.0 & 61.4
              & 67.5 & 72.3 & \textbf{75.2} & 62.1 & 69.3
              & 35.5 & 39.6 & 37.9 & 36.7 & 37.4
              & 40.0 & 47.0 & 45.1 & 43.9 & 44.1 \\
BAM ~\cite{lang2022learning}           & 
              & 69.0 & 73.6 & 67.6 & 61.1 & 67.8
              & 70.5 & 75.1 & 70.8 & 67.2 & 70.9
              & 39.4 & 49.9 & 46.2 & 45.2 & 45.2
              & 43.2 & 53.4 & 49.4 & 48.1 & 48.5 \\
HDMNet~\cite{peng2023hierarchical}         & 
              & 71.0 & 75.4 & 68.9 & 62.1 & 69.4
              & 71.3 & 76.2 & 71.3 & \textbf{68.5} & 71.8
              & 43.8 & 55.3 & 51.6 & 49.4 & 50.0
              & 50.6 & 61.6 & 55.7 & 56.0 & 56.0 \\
VRP-SAM$^\dag$~\cite{sun2024vrp} & 
              & 74.5 & 77.3 & 69.5 & 65.8 & 71.8
              & 76.3 & 76.6 & 69.5 & 63.1 & 71.4
              & 44.3 & 54.3 & 52.3 & 50.0 & 50.2
              & 50.5 & 59.5 & 56.9 & 54.9 & 55.5 \\
HMNet~\cite{xu2024hybrid} &
              &72.2 & 75.4 & 70.0 & 63.9 & 70.4 &
              74.2 & 77.3 & 74.1 & 70.9 & 74.1 &
              45.5 & 58.7 & 52.9 & 51.4 & 52.1 &
              53.4 & 64.6 & 60.8 & 56.8 & 58.9 \\
UniFSS~\cite{chang2025beyond} &
              &72.7 & 75.6 & 63.7 & 66.9 & 69.7 &
              75.4 & 77.1 & 67.9 & 69.9 & 72.6 &
              46.5 & 53.0 & 48.0 & 48.2 & 48.9 &
              50.3 & 59.5 & 54.4 & 52.0 & 54.1 \\
FCP$^\dag$~\cite{park2025foreground}    & 
              & 74.9 & 77.4 & 71.8 & 68.8 & 73.2
              & 77.2 & 78.8 & 72.2 & 67.7 & 74.0
              & 46.4 & 56.4 & 55.3 & 51.8 & 52.5
              & 52.6 & 63.3 & 59.8 & 56.1 & 58.0 \\
\rowcolor{gray!15}
              VINE (Ours)     & 
              & \textbf{76.5} & \textbf{78.7} & \textbf{72.6} & \textbf{69.1} & \textbf{74.2}
              & \textbf{78.6} & \textbf{80.9} & 72.8 & \textbf{68.2} & \textbf{75.1}
              & \textbf{47.7} & \textbf{57.4} & \textbf{56.7} & \textbf{52.9} & \textbf{53.7}
              & \textbf{54.1} & \textbf{64.6} & \textbf{61.3} & \textbf{57.4} & \textbf{59.3} \\
\hline
\end{tabular}
}
\caption{Comparison with state-of-the-art methods on PASCAL-5$^i$ and COCO-20$^i$ datasets under 1-shot and 5-shot settings. The mean mIoU (\%) across 4 folds (F-0 to F-3) is reported. † denotes methods based on SAM.}
\label{tab:main_results}
\vspace{-1.0em}
\end{table*}

\begin{table}[t]
  \vspace{-0.5em}
  \centering
  \small
  \setlength{\tabcolsep}{5pt}
  \renewcommand{\arraystretch}{1.05}
  \resizebox{\linewidth}{!}{%
    \begin{tabular}{l|c|c|c|c}
      \toprule
      Method & Train (A) & Test (B) & mIoU (\%) & $\Delta$ \\
      \midrule
      Baseline (similar)   & dog   & person    & 43.69 & -- \\
      Ours                 & dog   & person    & \textbf{53.41} & +9.72 \\
      \midrule
      Baseline (typical)   & horse & person    & 44.28 & -- \\
      Ours                 & horse & person    & \textbf{52.62} & +8.34 \\
      \midrule
      Baseline (divergent) & dog   & motorbike & 17.76 & -- \\
      Ours                 & dog   & motorbike & \textbf{36.28} & +18.52 \\
      \bottomrule
    \end{tabular}%
  }
  \vspace{-0.3em}
  \caption{Cross-class generalization on PASCAL-5$^i$ (fold-2). }
  \vspace{-1.5em}
  \label{tab:cross-domain}
\end{table}
\section{Experiments}
\label{sec:experiments}
% image
% image
\subsection{Implemetation Details}
\paragraph{Datasets.}
We evaluate our method on two widely adopted few-shot segmentation benchmarks: PASCAL-5$^i$~\cite{shaban2017one} and COCO-20$^i$~\cite{nguyen2019feature}. 
PASCAL-5$^i$ contains 20 object categories selected from PASCAL VOC 2012~\cite{everingham2010pascal} and SDS~\cite{hariharan2014simultaneous}, while COCO-20$^i$ includes 80 categories from MS COCO~\cite{lin2014microsoft}. 
Both datasets are divided into 4 disjoint folds without class overlap. 
Following the standard evaluation protocol, we adopt a 3-fold training and 1-fold testing scheme, sampling 1,000 support-query pairs per test fold from novel classes. 
We report mean Intersection-over-Union (mIoU) as the primary quantitative evaluation metric.
\paragraph{Experimental Setup.}
We adopt FCP~\cite{park2025foreground} as the baseline. Both the SAM encoder and auxiliary visual backbones (VGG-16, ResNet-50) remain frozen during training. Models are optimized with AdamW under a cosine-annealing schedule. For PASCAL-5$^i$, training runs 100 epochs at a learning rate of $2\times10^{-4}$, and for COCO-20$^i$, 50 epochs at $1\times10^{-4}$. The input resolution is $512{\times}512$ with a batch size of 8. Each support and query branch is initialized with 50 learnable prompt tokens. For pseudo-view generation, the perspective perturbation parameter $\delta_{\max}$ is set to 0.001 to control geometric variation.

\subsection{Results and Analysis}
\subsubsection{Comparison with State-of-the-Art Methods}
We evaluate the proposed \textbf{VINE} framework on two standard few-shot segmentation benchmarks: PASCAL-5$^{i}$ and COCO-20$^{i}$. Comparisons are made against state-of-the-art prompt-based approaches as well as representative prototype-matching methods. 
As summarized in Table~\ref{tab:main_results}, VINE delivers consistently strong results across backbones (VGG-16 and ResNet-50), shot configurations (1-shot and 5-shot), and datasets.
On PASCAL-5$^{i}$, VINE achieves 74.2\% and 75.1\% mIoU in the 1-shot and 5-shot settings, surpassing the strongest baseline FCP by +2.1 and +1.1 points, respectively. 
On COCO-20$^{i}$, which exhibits higher intra-class variation and more cluttered backgrounds, VINE attains 53.7\% and 59.3\% mIoU, improving upon FCP by +2.0 and +1.3 points.
These gains, observed consistently across both benchmarks, validate the effectiveness of the proposed view-consistent prototype learning.
\subsubsection{Cross-class prototype stability analysis}
\begin{table}[t]
  \vspace{-0.4em}
  \centering
  \small
  \setlength{\tabcolsep}{7pt}
  \renewcommand{\arraystretch}{1.1}
  \resizebox{\linewidth}{!}{
    \begin{tabular}{c|cccc|c|c}
      \toprule
      \textbf{ID} & \textbf{Res} & \textbf{SAM} & \textbf{DFM} & \textbf{SVGA} & \textbf{mIoU (\%)} & \boldmath$\Delta$ \\
      \midrule
      (a) & \checkmark &            &            &            & 71.8          & --   \\
      (b) &            & \checkmark &            &            & 66.2          & -5.6 \\
      (c) & \checkmark & \checkmark &            &            & 72.5          & +0.7 \\
      (d) & \checkmark & \checkmark & \checkmark &            & 73.3          & +1.5 \\
      (e) & \checkmark & \checkmark &            & \checkmark & 73.8          & +2.0 \\
      (f) & \checkmark & \checkmark & \checkmark & \checkmark & \textbf{74.2} & +2.4 \\
      \bottomrule
    \end{tabular}
  }
  \vspace{-0.3em}
  \caption{Ablation study of key components in VINE. }
  \label{tab:key-component}
  \vspace{-1.2em}
\end{table}
To assess prototype stability under distribution shift on PASCAL-5\textsuperscript{i}, we use cross-class matching (support and query from disjoint categories), following FCP~\cite{park2025foreground}. This isolates the model’s ability to recover structural cues without class-level semantics. Table~\ref{tab:cross-domain} summarizes three regimes. In the \emph{similar} case (dog→person), shared textures cause FCP to activate on background edges; our discriminative foreground modulation suppresses these distractors and improves boundary reliability (+9.72 mIoU). In the \emph{typical} case (horse→person), viewpoint and pose changes destabilize FCP’s prototype alignment; SVGA propagates view-consistent cues across instances, restoring geometric coherence (+8.34 mIoU). In the \emph{divergent} case (dog→motorbike), with no semantic or structural commonality, FCP collapses (17.76 mIoU). Our full model preserves spatial coherence through structural alignment and foreground emphasis (+18.52 mIoU). As semantic and geometric disparity increases, our method maintains stable prototype formation, confirming the need for cross-view structural reasoning and discriminative modulation under cross-class conditions.
\begin{figure}[t]
\centering

% ---- Left figure: pseudo-mask quality ----
\begin{minipage}{0.48\linewidth}
    \centering
    \includegraphics[width=1.1\linewidth]{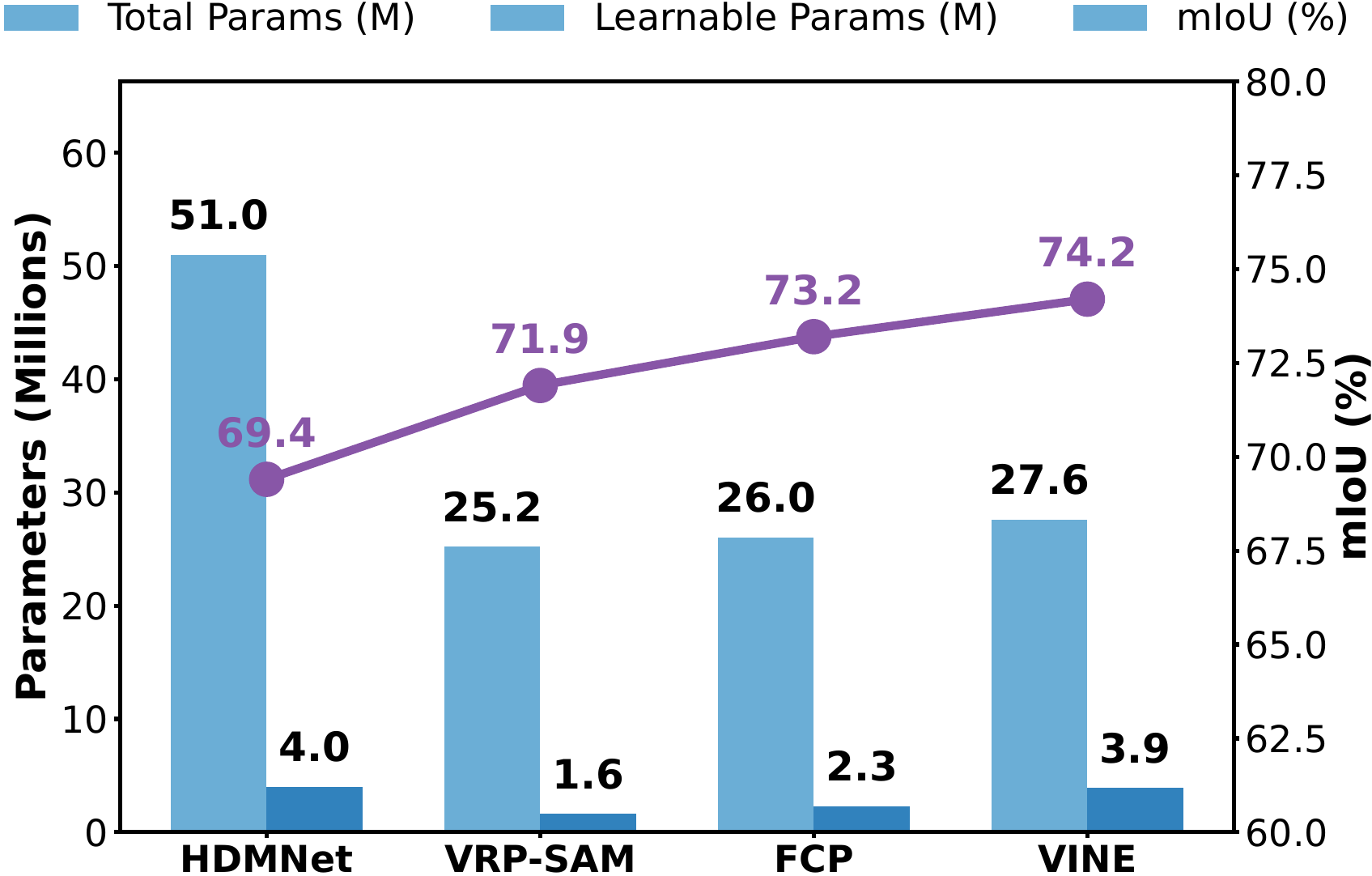}
    \vspace{-1.75em}
    \caption{Parameter efficiency Comparison.}
    \label{fig:params_miou_efficiency}
\end{minipage}
\hfill
% ---- Right figure: parameter efficiency + mIoU ----
\begin{minipage}{0.47\linewidth}
    \centering
    \includegraphics[width=1.0\linewidth]{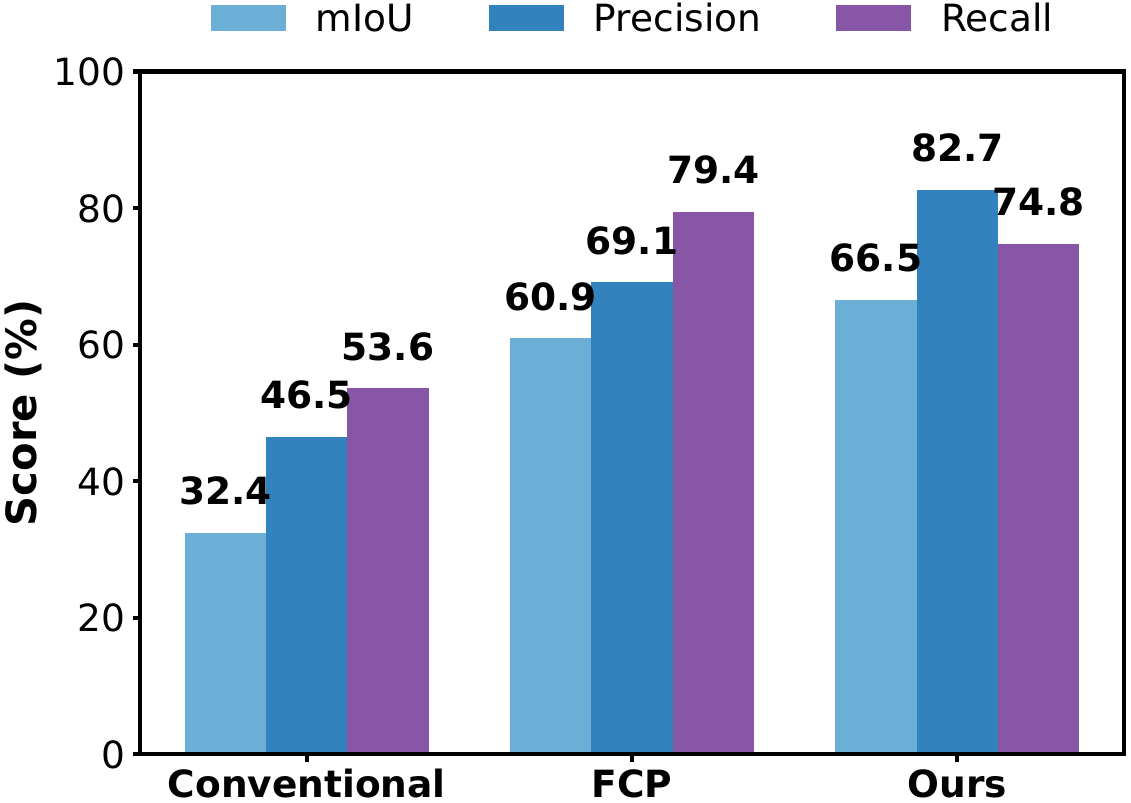}
    \vspace{-1.75em}
    \caption{Pseudo-mask quality under different methods.}
    \label{fig:pseudo_mask_quality}
\end{minipage}

\vspace{-1.6em}
\end{figure}

\subsection{Ablation Study} 
\paragraph{Main Components.}
We conduct ablation studies to assess the contribution of each module, summarized in Table~\ref{tab:key-component}. Using ResNet-50 alone (a) serves as the baseline. Replacing it with the SAM encoder (b) leads to a 5.6\% drop in mIoU, indicating that semantic features without structural priors reduce spatial precision. Adding SAM alongside ResNet-50 (c) brings a modest +0.7\% gain. Incorporating DFM (d) further improves foreground discrimination by modeling support--query discrepancies. Substituting DFM with SVGA (e) instead yields +2.0\% through dual-graph reasoning that captures spatial relations and cross-view correspondence. To further analyze SVGA module, we decompose it into spatial and view graphs (Table~\ref{tab:ablations_all}(b)): the spatial graph enhances local geometry, while the view graph contributes larger gains by enforcing cross-view consistency. Their combination achieves the best result, validating complementary intra-view detail and inter-view alignment. Finally, integrating DFM and SVGA (f) reaches 74.2\% (+2.4\%), confirming their synergy in refining prototypes under structural and appearance variations.
\paragraph{Loss Function Analysis.} 
Table~\ref{tab:ablations_all} summarizes the loss composition and weighting. 
Using only $\mathcal{L}_{\text{predict}}$ provides pixel supervision but no structural regularization, whereas using only $\mathcal{L}_{\text{proto}}$ yields a small gain (+0.2\%) by enforcing support–query alignment. 
Combining both is best (+0.8\%), showing their complementarity: $\mathcal{L}_{\text{predict}}$ ensures pixel fidelity and $\mathcal{L}_{\text{proto}}$ preserves prototype coherence. 
Sensitivity analysis indicates that $\lambda_{\text{proto}}{=}1.0$, $\lambda_{\text{predict}}{=}0.5$ performs best; larger $\lambda_{\text{predict}}$ weakens structural alignment, and smaller values hurt localization. 
Thus, joint optimization of local prediction and global prototype alignment is key to robust FSS under cross-view variation.
\paragraph{Effectiveness of Discriminative Pseudo Masks.}
As shown in Fig.~\ref{fig:pseudo_mask_quality}, the quality of pseudo masks differs markedly across methods. 
Similarity-based masks exhibit low precision under background clutter, while FCP~\cite{park2025foreground}, though benefiting from attention cues, remains susceptible to over-activation along object boundaries. 
Our DFM module produces substantially cleaner and more discriminative pseudo masks by exploiting support--query discrepancies to construct a query-adaptive saliency prior. 
This leads to a notable improvement in mIoU and a pronounced gain in precision, indicating more reliable foreground delineation and reduced boundary leakage. 
The resulting prototypes show greater structural integrity, supporting more stable query localization under appearance changes.
\begin{table}[t]
\centering
\small
\renewcommand{\arraystretch}{0.9}
\setlength{\tabcolsep}{18pt}

\begin{tabular}{l|c|c}
\toprule
\textbf{Configuration} & \textbf{mIoU (\%)} & \textbf{$\Delta$} \\
\midrule

\rowcolor{gray!15}
\multicolumn{3}{l}{\textbf{(1) Loss components}} \\

\hspace{1em} (a) Only $\mathcal{L}_{\text{predict}}$     & 73.4 & -- \\
\hspace{1em} (b) Only $\mathcal{L}_{\text{proto}}$        & 73.6 & +0.2 \\
\hspace{1em} (c) Both losses                               & \textbf{74.2} & +0.8 \\
\midrule

\rowcolor{gray!15}
\multicolumn{3}{l}{\textbf{(2) SVGA graph configuration}} \\

\hspace{1em} None                 & 73.3 & -- \\
\hspace{1em} Spatial only         & 73.4 & +0.1 \\
\hspace{1em} View only            & 74.0 & +0.7 \\
\hspace{1em} Both graphs          & \textbf{74.2} & +0.9 \\
\midrule

\rowcolor{gray!15}
\multicolumn{3}{l}{\textbf{(3) Loss-weight sensitivity $\lambda_{\text{proto}}$ / $\lambda_{\text{predict}}$}} \\

\hspace{1em} 1.0 / 0.3   & 72.2 & -- \\
\hspace{1em} \textbf{1.0 / 0.5}  & \textbf{74.2} & +2.0 \\
\hspace{1em} 1.0 / 1.0   & 72.5 & +0.3 \\
\hspace{1em} 0.5 / 1.0   & 73.3 & +1.1 \\
\bottomrule
\end{tabular}
\caption{Ablation studies on key hyper-parameters. The analysis includes (1) the impact of loss composition, (2) the configuration of spatial and view graphs in SVGA, and (3) the sensitivity to loss-weight ratios $(\lambda_{\text{proto}}, \lambda_{\text{predict}})$.}
\label{tab:ablations_all}
\vspace{-1.5em}
\end{table}
\begin{figure*}[t]
    \centering
    % 左子图
    \begin{subfigure}[t]{0.575\linewidth}
        \centering
        \includegraphics[width=\linewidth]{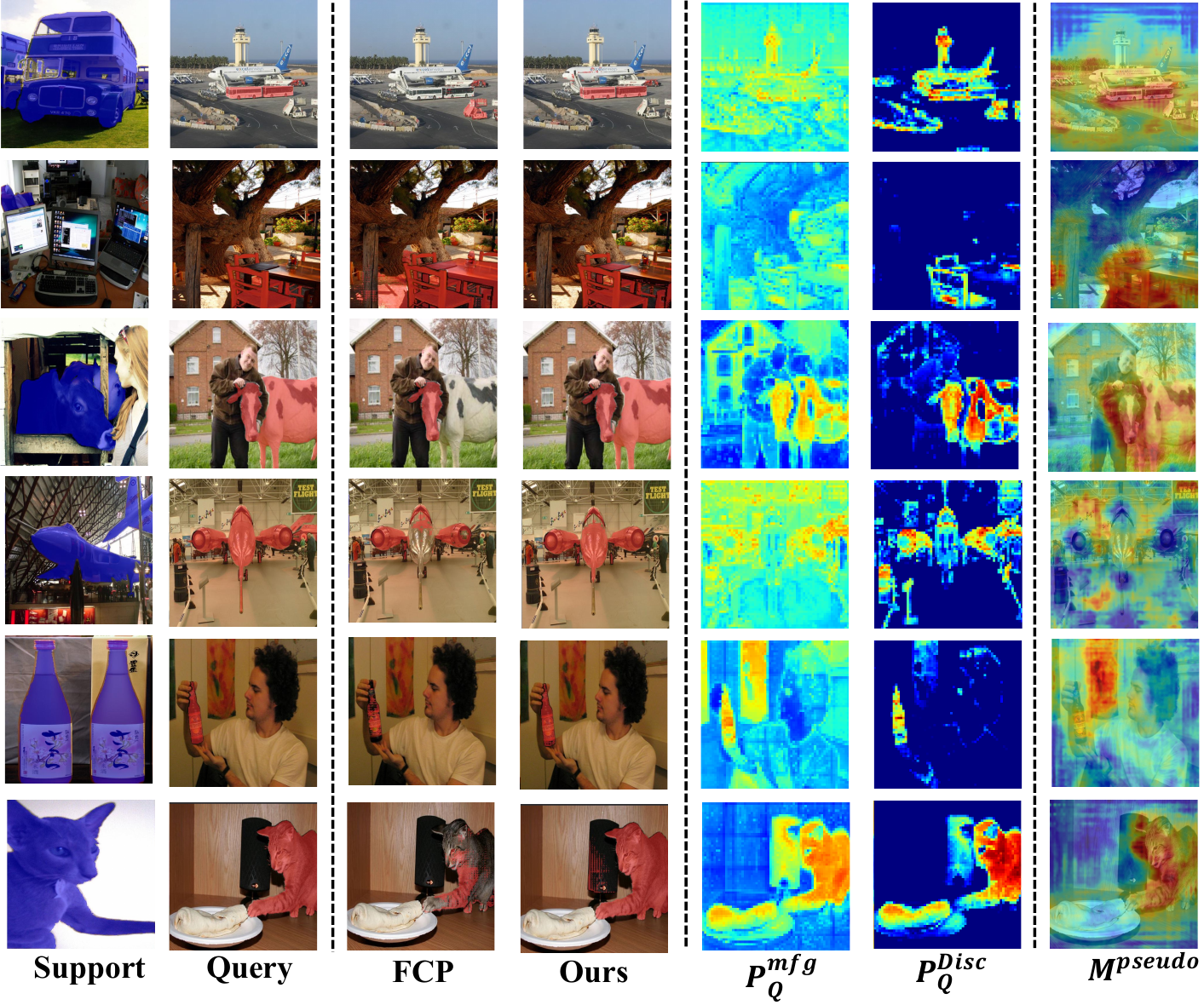}
        \caption{Qualitative comparison between FCP and VINE on representative cross-view query-support image pairs from the PASCAL-5$^i$ dataset.}
        \label{fig:qualitative}
    \end{subfigure}
    \hspace{0.01\linewidth}
   % 右图 - 手动上提对齐
    \begin{subfigure}[t]{0.39\linewidth}
        \centering
        \raisebox{4.5mm}{  % 可调数值，3mm～6mm之间微调
        \includegraphics[width=\linewidth]{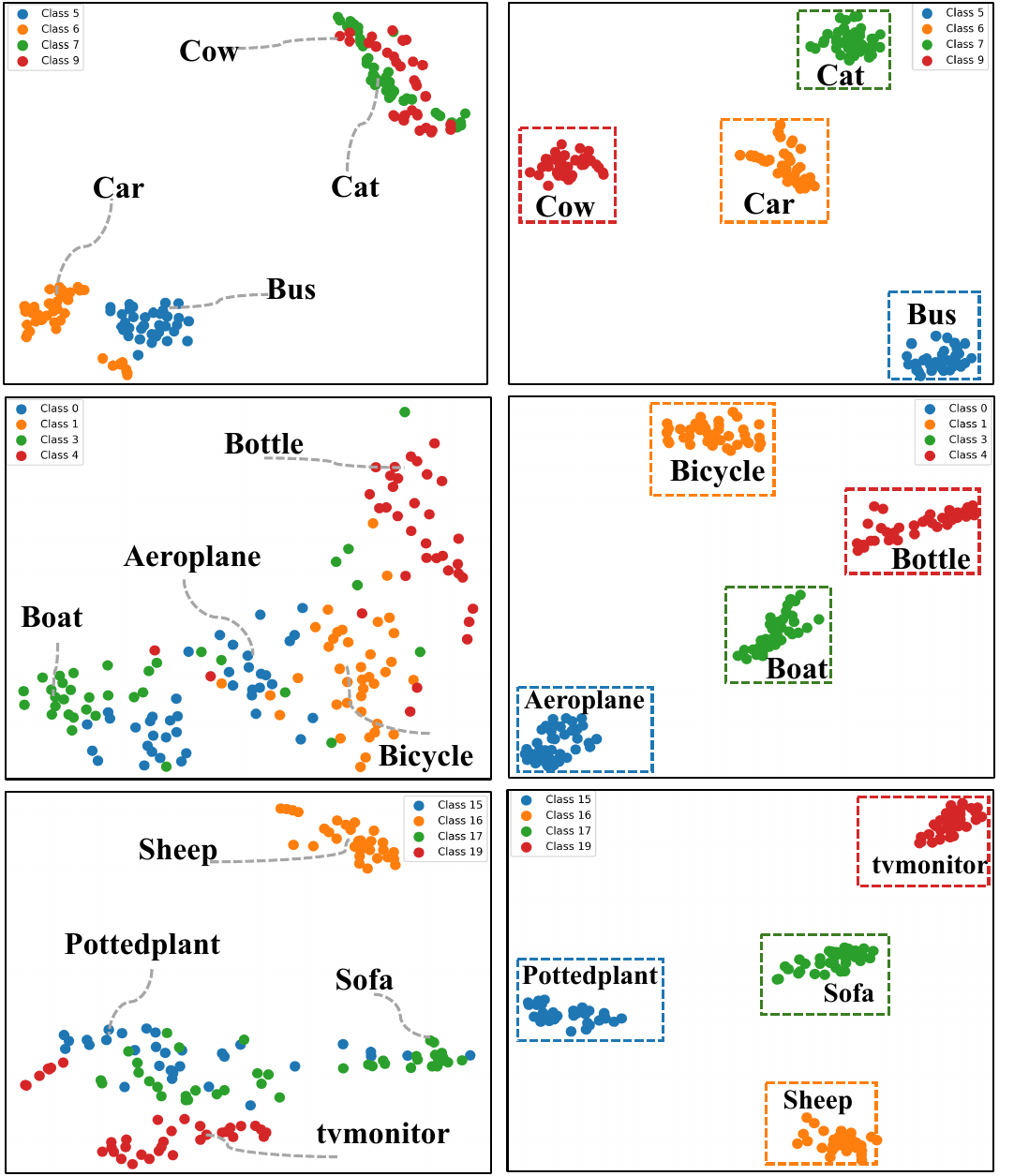}
        }
        \caption{t-SNE visualization of query feature distributions on PASCAL-5$^i$ folds 0,1 and 3.}
        \label{fig:tsne_svga}
    \end{subfigure}
    \vspace{-0.5em}
    \caption{Qualitative and feature-level comparison of segmentation results.}
    \vspace{-1.5em}
    \label{fig:qual_+_tsne}
\end{figure*}
\paragraph{Parameter Study on Model Efficiency.}
Fig.~\ref{fig:params_miou_efficiency} compares model capacity and segmentation accuracy across representative prompt-based frameworks, emphasizing the detailed comparison with FCP~\cite{park2025foreground}. 
VINE attains the highest mIoU of 74.2\% with only 27.6M parameters, surpassing FCP (73.2\%, 26.0M) by +1.0\% with merely a 6\% increase in total size. 
This small growth contrasts with substantial quality gains, clearly showing that improvements stem not from scaling but from view-consistent prototype formulation and discriminative foreground modulation. 
Despite similar learnable parameters (3.9M vs. 2.3M), VINE achieves higher efficiency in translating capacity into accuracy. 
These results further confirm the effectiveness of our structural alignment design, balancing accuracy and compactness for robust few-shot generalization.
% 多个参数和消融实验
% \begin{table}[t]
% \centering
% \small
% \renewcommand{\arraystretch}{1.15}

% % ====================================================
% % 固定表格整体宽度
% \begin{tabularx}{\linewidth}{@{} c X X X X @{}}
% % ====================================================

% % --------- Block 1 Header ---------
% \rowcolor{gray!15}
% \multicolumn{5}{c}{\textbf{Loss components} (\,$\mathcal{L}_{\text{predict}}$ / $\mathcal{L}_{\text{proto}}$\,)}\\
% \toprule

% % --------- Block 1 Rows (4 columns, 5th empty for alignment) ---------
% Config 
% & (a) $\checkmark$/-- 
% & (b) --/$\checkmark$ 
% & (c) $\checkmark$/$\checkmark$ 
% & \multicolumn{1}{c}{} \\

% mIoU   
% & 73.4 
% & 73.6 
% & \textbf{74.2} 
% & \multicolumn{1}{c}{} \\
% \midrule

% % --------- Block 2 Header ---------
% \rowcolor{gray!15}
% \multicolumn{5}{c}{\textbf{SVGA graph configurations} (Spatial / View)}\\
% \midrule

% % --------- Block 2 Rows ---------
% Config 
% & $\times$/$\times$ 
% & $\checkmark$/$\times$
% & $\times$/$\checkmark$
% & $\checkmark$/$\checkmark$ \\

% mIoU   
% & 73.3 
% & 73.4 
% & 74.0 
% & \textbf{74.2} \\
% \midrule

% % --------- Block 3 Header ---------
% \rowcolor{gray!15}
% \multicolumn{5}{c}{\textbf{Loss weight sensitivity} (\,$\lambda_{\text{proto}}$ / $\lambda_{\text{predict}}$\,)}\\
% \midrule

% % --------- Block 3 Rows ---------
% Setting
% & 1.0/0.3
% & \textbf{1.0/0.5}
% & 1.0/1.0
% & 0.5/1.0 \\

% mIoU
% & 72.2
% & \textbf{74.2}
% & 72.5
% & 73.3 \\
% \bottomrule
% \end{tabularx}

% \caption{Ablations over loss components, SVGA configurations, and loss-weight sensitivity.}
% \label{tab:ablations_all}
% \end{table}
\subsection{Visualization Results}
\paragraph{Qualitative results.} 
Fig.~\ref{fig:qualitative} visualizes representative query–support pairs together with three cues—the semantic prior $P_{Q,fg}^{m}$, the similarity-based pseudo mask $M^{\text{pseudo}}$, and our discriminative prior $P_{Q}^{\text{Disc}}$.
Under cross-view mismatch (frontal vs.\ lateral buses; upward vs.\ frontal airplanes), the FCP baseline emphasizes locally salient parts and breaks geometric continuity, producing fragmented masks; SVGA propagates structural relations across viewpoints and restores object extent.
When the support provides only partial semantics (cow head; half a cat body), attention in baseline remains part-biased and fails to recover the full object; combining SVGA with DFM yields a query-adaptive foreground prior that completes the object while suppressing background.
In cluttered scenes (small handheld bottles;obscured chairs), similarity-driven $M^{\text{pseudo}}$ introduces boundary noise and leakage, whereas the joint SVGA–DFM modeling maintains compact, structurally consistent, and semantically focused prototypes.
Across cases, $P_{Q,fg}^{m}$ offers coarse localization, $M^{\text{pseudo}}$ degrades under view changes and clutter, and $P_{Q}^{\text{Disc}}$ produces sharper, spatially aware activations—mirroring the quantitative gains in Section~\ref{sec:experiments} (more examples in the Appendix).
\vspace{-1.0em}
\paragraph{t-SNE Visualization.}
As shown in Fig.~\ref{fig:tsne_svga}, SVGA markedly improves the \emph{feature geometry} under viewpoint variation. 
Baseline embeddings exhibit scattered intra-class patterns and mixed inter-class regions---e.g., ``Cow'' vs.\ ``Cat'' (fold 0) and ``Aeroplane'', ``Bicycle'', ``Boat'' (fold 1)---reflecting deformation sensitivity and appearance bias. 
With SVGA, features become compact and well separated, indicating stronger structural alignment and semantic disentanglement. 
Previously fragmented categories such as ``Aeroplane'', ``Boat'', and ``Tvmonitor'' form cohesive clusters with sharper margins, showing that dual-graph reasoning effectively propagates geometric priors across views. 
Overall, SVGA enhances intra-class compactness and inter-class separability, yielding a stable and view-consistent space for prototype formation in FSS.
\section{Conclusion}
\label{sec:conclusion}
In this paper, we present VINE, a unified framework for few-shot segmentation that jointly enforces structural consistency and foreground discrimination. VINE constructs spatial and view graphs to capture geometry and viewpoint variation, while deriving query-adaptive foreground priors from support–query differences. The resulting class-consistent prototypes function as reliable visual reference cues to guide mask prediction. Extensive experiments show that VINE consequently improves segmentation robustness under appearance shifts and view changes, offering a principled and versatile solution with strong generalization.

\section{Acknowledgements}
\label{sec:Acknowledgements}
This work was supported in part by the National Key Research and Development Project under Grant 2023YFC3806000, in part by the National Natural Science Foundation of China under Grant 62406226, in part sponsored by Shanghai Sailing Program under Grant 24YF2748700, in part by New-Generation Information Technology under the Shanghai Key Technology R\&D Program under Grant 25511103500.
{
    \small
    \bibliographystyle{ieeenat_fullname}
    \bibliography{main}
}


\end{document}